\documentclass[10pt,twocolumn,letterpaper]{article}

\usepackage{packages/cvpr}
\usepackage{times}
\usepackage{epsfig}
\usepackage{graphicx}
\usepackage{amsmath}
\usepackage{amssymb}
\usepackage{booktabs}
\usepackage{subfig}

\usepackage{enumitem}
\usepackage{bm}
\usepackage{packages/algorithm}
\usepackage{packages/algorithmic}\usepackage{array} \newcolumntype{x}[1]{>{\centering\let\newline\\\arraybackslash\hspace{0pt}}m{#1}}
\newcommand\scalemath[2]{\scalebox{#1}{\mbox{\ensuremath{\displaystyle #2}}}}

\usepackage{array}
\newcommand{\PreserveBackslash}[1]{\let\temp=\\#1\let\\=\temp}
\newcolumntype{C}[1]{>{\PreserveBackslash\centering}m{#1}}
\newcolumntype{R}[1]{>{\PreserveBackslash\raggedleft}p{#1}}
\newcolumntype{L}[1]{>{\PreserveBackslash\raggedright}p{#1}}

\cvprfinalcopy 
\def\cvprPaperID{6223} 

\ifcvprfinal\fi
\begin{document}

\title{An Efficient Schmidt-EKF for 3D Visual-Inertial SLAM}

\author{Patrick Geneva, \quad James Maley, \quad Guoquan Huang\\
University of Delaware\\
{\tt\small pgeneva@udel.edu, james.m.maley2.civ@mail.mil, ghuang@udel.edu}
}

\maketitle
\ifcvprfinal\thispagestyle{empty}\fi

\begin{abstract}

It holds great implications for practical applications to enable centimeter-accuracy positioning for mobile and wearable sensor systems.
In this paper, we propose a novel, high-precision, efficient visual-inertial (VI)-SLAM algorithm, termed Schmidt-EKF VI-SLAM (SEVIS),
which optimally fuses IMU measurements and monocular images in a tightly-coupled manner to provide 3D motion tracking with bounded error.
In particular, we adapt the Schmidt Kalman filter formulation to selectively include informative features in the state vector while treating them as nuisance parameters (or Schmidt states) once they become matured.
This change in modeling allows for significant computational savings by no longer needing to constantly update the Schmidt states (or their covariance), while still allowing the EKF to correctly account for their cross-correlations with the active states.
As a result, we achieve linear computational complexity in terms of map size,  instead of quadratic as in the standard SLAM systems.
In order to fully exploit the map information to bound navigation drifts,
we advocate efficient keyframe-aided 2D-to-2D feature matching to find reliable correspondences between current 2D visual measurements and 3D map features.
The proposed SEVIS is extensively validated in both simulations and experiments.

\end{abstract}

\section{Introduction}

Enabling {centimeter-accuracy positioning}  for mobile and wearable devices such as smart phones and micro air vehicles (MAVs), holds potentially huge implications for practical applications.
One of the most promising methods providing precision navigation in 3D
is through the fusion of visual and inertial sensor measurements (i.e., visual-inertial navigation systems or VINS)~\cite{Mourikis2007ICRA,Hesch2013TRO,Li2013IJRR,Leutenegger2014IJRR,Huang2014ICRA,Huai2018IROS}.
This localization solution has the advantages of being both cheap and ubiquitous, 
and has the potential to provide position and orientation (pose) estimates which are on-par in terms of accuracy with more expensive sensors such as LiDAR. 
To date, various algorithms are available for VINS problems 
including visual-inertial (VI)-SLAM~\cite{Kim2007RAS,Shen2015ICRA}
and visual-inertial odometry (VIO)~\cite{Mourikis2007ICRA,Mourikis2009TRO,Li2013IJRR}, 
such as the extended Kalman filter (EKF)~\cite{Mourikis2007ICRA,Kottas2012ISER,Hesch2014IJRR,Li2013IJRR,Huang2014ICRA,Huang2015ISRR,Wu2015RSS,Paul2017ICRA},
unscented Kalman filter (UKF)~\cite{Ebcin2007TR,Brossard2018FUSION}, 
and batch or sliding-window optimization methods~\cite{Strelow2004,Indelman2013RAS,Leutenegger2014IJRR,Mur-Artal2017RAL,Yang2017TASE,Shen2015ICRA,Qin2018TRO},
among which the EKF-based approaches remain arguably the most popular for resource constrained devices because of their efficiency.
While current approaches can perform well over a short period of time in a small-scale environment
(e.g., see \cite{Hesch2013TRO,Li2013IJRR,Huai2018IROS}),
they are not robust and accurate enough for long-term, large-scale deployments in challenging environments,
due to their limited available resources of sensing, memory and computation, 
which, if not properly addressed, often result in short mission duration or intractable real-time estimator performance.

In this paper, we will primarily focus on EKF-based VI-SLAM rather than VIO.  
VI-SLAM has the advantage of building a map of the surrounding environment, which enables ``loop closing'' to bound long-term navigation drift.
VIO systems do not build a map and therefore cannot leverage information from prior observations to help improve estimator performance.
However, one of the largest downsides of SLAM is that its computational complexity grows quadratically with the number of landmarks in the map, which commonly makes it computationally intractable without simplifying assumptions to allow for them to run on resource constrained sensor platforms such as mobile devices.
To address this complexity issue, we leverage the computationally-efficient multi-state constraint Kalman filter (MSCKF)~\cite{Mourikis2007ICRA}
and selectively keep a number of features (say $n$) in the state vector as a map of the environment,
enabling the system to use them for a long period of time and thus allowing for (implicit) loop closures to bound drifts.
This, however, would still exhibit ${O}(n^2)$ computational complexity as in the standard EKF-based SLAM.
By observing that features' estimates do not have significant updates if they approach their steady state (i.e., becoming matured/converged),
we could gain substantial computational savings by avoiding performing EKF updates for those matured map features while still taking into account their uncertainty.
To this end, we adapt the Schmidt Kalman filter (SKF) \cite{Schmidt1966ACS} and treat map features as nuisance parameters 
which will no longer be updated but whose covariance and cross-correlations to other states are still utilized in the EKF update.
As a result, this renders only ${O}(n)$ computational complexity, 
making our proposed {\bf S}chmidt-{\bf E}KF {\bf V}isual-{\bf I}nertial {\bf S}LAM (SEVIS) significantly more amenable to running on resource-constrained sensor platforms.

In particular, the main contributions of the paper include:
\begin{itemize}[nolistsep]
    \item We design a high-precision, efficient Schmidt-EKF based VI-SLAM (i.e., SEVIS) algorithm 
    which leverages the Schmidt-KF formulation to allow for concurrent estimation of an environmental map used for long-term loop closures to bound navigation drifts with linear computational complexity.

    \item  We propose a keyframe-aided 2D-to-2D matching scheme for the challenging data association problem of matching 2D visual measurements to 3D map features, 
    without performing 3D-to-2D matching (which may not be applicable to sparse 3D environmental maps).
    This 2D-to-2D matching is not effected by estimation performance, allowing for long-term loop closures and recovery from extreme drifts.

    \item We validate the proposed SEVIS algorithm extensively in both Monte-Carlo simulations and real-world experiments, showing the applicability and performance gains offered by our system. 
    The experimental study of computation requirements further shows that the proposed SEVIS remains real-time while building and maintaining a 3D feature-based map.
\end{itemize}

 \section{Related Work}

While SLAM estimators -- by jointly estimating the location of the sensor platform and the features in the surrounding environment --
are able to easily incorporate loop closure constraints to bound localization errors and have attracted much research attention in the past three decades~\cite{Durrant-Whyte2006RAM,Bailey2006RAM,Cadena2016TRO,Bresson2017TIV},
there are also significant research efforts devoted to open-loop VIO systems  (e.g.,~\cite{Mourikis2007ICRA,Hesch2013TRO,Hesch2014IJRR,Li2013IJRR,Huang2014ICRA,Wu2015RSS,Paul2017ICRA,Wu2017IROS,Zhang2017RAL,Brossard2017IROS,Bloesch2017IJRR,Huai2018IROS,Qin2018TRO}).
For example, a hybrid MSCKF/SLAM estimator was developed for VIO~\cite{Li2012RSS},
which retains features that can be continuously tracked beyond the sliding window in the state as SLAM features while removing them when they get lost.

It is challenging to achieve accurate localization by performing large-scale VI-SLAM 
due to the inability to remain computationally efficient 
without simplifying assumptions such as treating keyframe poses and/or map features to be perfect (i.e., zero uncertainty).
Many methods use feature observations from different keyframes to limit drift over the trajectory (e.g., \cite{Nerurkar2014ICRA,Leutenegger2014IJRR}),
and with most leveraging a two-thread architecture that optimizes a small window of local keyframes and features, while a background thread solves a long-term sparse pose graph containing loop closure constraints~\cite{Engel2014ECCV,Mur2017RAL,Liu2018CVPR,Qin2018TRO,Qin2018RELOC}.
For example, VINS-Mono~\cite{Qin2018TRO,Qin2018RELOC} uses loop closure constraints 
in both the local sliding window and in the global batch optimization.
During the local optimization, feature observations from keyframes provide implicit loop closure constraints, 
while the problem size remains small by assuming the keyframe poses are perfect (thus removing them from optimization), while their global batch process optimizes a relative pose graph.
In \cite{Mourikis2008CVPRW} a dual-layer estimator uses the MSCKF to perform real-time motion tracking
and triggers the global bundle adjustment (BA) on loop closure detection.
This allows for the relinearization and inclusion of loop closure constraints in a consistent manner, while requiring substantial additional overhead time where the filter waits for the BA to finish.
A large-scale map-based VINS~\cite{Lynen2015RSS} assumes a compressed prior map containing feature positions and their uncertainty and uses matches to features in the prior map to constrain the localization globally.
The recent Cholesky-Schmidt-KF~\cite{Dutoit2017ICRA} however 
explicitly considers the uncertainty of the prior map, by employing the sparse Cholesky factor of the map's information matrix and further relaxing it by reducing the map size with more sub-maps for efficiency.
In contrast, in this work, we formulate a single-threaded Schimdt-EKF for VI-SLAM, allowing for full probabilistic fusion of measurements without sacrificing real-time performance
and permitting the construction and leverage of an environmental map to bound long-term navigation drift indefinitely.

 \section{Visual-Inertial SLAM} \label{sec:vi-slam}

The process of VI-SLAM optimally fuses camera images and IMU (gyroscope and accelerometer) measurements to provide 6DOF pose estimates of the sensor platform as well as reconstruct 3D positions of environmental features (map).
In this section, we briefly describe VI-SLAM within the EKF framework, which serves as the basis for our proposed SEVIS algorithm.

The state vector of VI-SLAM contains the IMU navigation state $\mathbf{x}_I$ and a sliding window of cloned past IMU (or camera) poses $\mathbf{x}_C$ as in the MSCKF \cite{Mourikis2007ICRA}, 
as well as the map features' positions $\mathbf{x}_S$ expressed in the global frame:\footnote{Throughout this paper 
the subscript $\ell |j$ refers to the estimate of a
quantity at time-step $\ell$, after all measurements up to time-step $j$ have been processed. $\hat
x$ is used to denote the estimate of a random variable $x$, while $\tilde x = x-\hat x$ is  the error in this estimate. 
$\mathbf I_{n\times m}$ and $\mathbf 0_{n\times m}$ are the $n \times m$ identity and zero matrices, respectively.
Finally, the left superscript denotes the frame of reference the vector is expressed with respect to.} 
\begin{align} 
\mathbf{x}_k &= \begin{bmatrix} \mathbf{x}_I^\top & \mathbf{x}_C^\top 
& \mathbf{x}_S^\top
\end{bmatrix}^{\top} =: \begin{bmatrix} \mathbf{x}_A^\top & \mathbf{x}_S^\top
\end{bmatrix}^{\top}
\label{eq:state}\\
\mathbf{x}_I &= \begin{bmatrix} {}_G^{I_{k}} \bar{q}{}^{\top} & \mathbf{b}_{\omega_k}^{\top} & {}^G\mathbf{v}_{{I}_{k}}^{\top} & \mathbf{b}_{a_k}^{\top} & {}^G\mathbf{p}_{{I}_{k}}^{\top}
\end{bmatrix}^{\top}  \\
\mathbf{x}_C &=
\scalemath{0.95}{\begin{bmatrix}
{}_G^{I_{k-1}} \bar{q}{}^{\top} &  {}^G\mathbf{p}_{I_{k-1}}^{\top} & 
\cdots & 
{}_G^{I_{k-m}} \bar{q}{}^{\top} & {}^G\mathbf{p}_{I_{k-m}}^{\top}
\end{bmatrix}^{\top}}
\\
\mathbf{x}_S &= 
\begin{bmatrix}
{}^G\mathbf{p}_{f_1}^{\top} & 
\cdots & 
{}^G\mathbf{p}_{f_n}^{\top}
\end{bmatrix}^{\top}
\end{align}
where ${}_G^{I_k}\bar{q}$ is the unit quaternion parameterizing the rotation $\mathbf{C}({}_G^{I_k}\bar{q})={}_G^{I_k} \mathbf{C}$ from the global frame of reference $\{G\}$ to the IMU local frame $\{I_k\}$ at time $k$ \cite{Trawny2005_Q_TR}, 
$\mathbf{b}_{\omega}$ and $\mathbf{b}_{a}$ are the gyroscope and accelerometer biases,
and ${}^G \mathbf{v}_{I_k}$ and ${}^G \mathbf{p}_{I_k}$ are the velocity and position of the IMU expressed in the global frame, respectively.
The clone state $\mathbf x_C$ contains $m$ historical IMU poses in a sliding window, while the map state $\mathbf{x}_S$ has $n$ features.
With the state decomposition~\eqref{eq:state}, the corresponding covariance matrix can be partitioned as:
\begin{align} \label{eq:cov}
                        \scalemath{0.9}{
    \mathbf{P}_k =
    \begin{bmatrix}
    \mathbf{P}_{AA_k} & \mathbf{P}_{AS_k} \\
    \mathbf{P}_{SA_k} & \mathbf{P}_{SS_k}
    \end{bmatrix} }
\end{align}

\subsection{IMU Propagation} \label{sec:propagation}

The inertial state $\mathbf{x}_I$ is propagated forward using incoming IMU measurements of linear accelerations ($\mathbf a_m$) and angular velocities ($\bm\omega_m$)
based on  the following generic nonlinear IMU kinematics~\cite{Chatfield1997}:
\begin{align}
    \mathbf{x}_{k+1} = \mathbf{f}(\mathbf{x}_{k}, \mathbf a_{m_k}-\mathbf n_{a_k}, \bm\omega_{m_k}-\mathbf{n}_{\omega_k} )
    \label{eq:imu_dynamics}
\end{align}
where $\mathbf n_a$ and $\mathbf n_\omega$ are the zero-mean white Gaussian noise of the IMU measurements.
We linearize this nonlinear model at the current estimate, and then propagate the state covariance matrix forward in time:
\begin{align} \label{eq:propcov}
    \scalemath{0.80}{
    \mathbf{P}_{k|k-1} =
    \begin{bmatrix}
    \bm\Phi_{k-1} \mathbf{P}_{AA_{k-1|k-1}}\bm\Phi_{k-1}^\top & \bm\Phi_{k-1} \mathbf{P}_{AS_{k-1|k-1}} \\
    \mathbf{P}_{SA_{k-1|k-1}}\bm\Phi_{k-1}^\top & \mathbf{P}_{SS_{k-1|k-1}} \end{bmatrix} 
    +
    \begin{bmatrix}
    \mathbf{Q}_{k-1} & \mathbf{0} \\
    \mathbf{0} & \mathbf{0}
    \end{bmatrix}} 
\end{align}
where $\bm\Phi_{k-1}$ and $\mathbf{Q}_{k-1}$ are respectively the system Jacobian and discrete noise covariance matrices for the active state \cite{Mourikis2007ICRA}.
Since the repeated computation of the above covariance propagation can become computationally intractable as the size of the covariance or rate of the IMU (e.g., $>$ 200Hz) grows,
we instead compound the state transition matrix and noise covariance as follows:
\begin{align}
    \bm\Phi(i+1) &= \bm\Phi_{k-1}\bm\Phi(i) \\
    \mathbf{Q}(i+1) &= \bm\Phi_{k-1}\mathbf{Q}(i)\bm\Phi_{k-1}^\top + \mathbf{Q}_{k-1}
\end{align}
with the initial conditions of $\bm\Phi(i=0)=\mathbf{I}$ and $\mathbf{Q}(i=0)=\mathbf{0}$.
After compounding $\bm\Phi(i+1)$ and $\mathbf{Q}(i+1)$, we directly apply them to propagate  $\mathbf{P}_{k-1|k-1}$ based on~\eqref{eq:propcov}.

\subsection{Camera Measurement Update}\label{sec:update}

Assuming a calibrated perspective camera, the measurement of a corner feature at time-step $k$ is 
the perspective projection of the 3D point, 
$^{C_k}\mathbf p_{f_i}$, expressed in the current camera frame $\{C_k\}$, 
onto the image plane, i.e.,
\begin{align} \label{eq:meas-model}
\mathbf z_{k} &= \frac{1}{z_k} \begin{bmatrix} x_k \\ y_k \end{bmatrix} + \mathbf n_{f_k} \\
\scalemath{.9}{\begin{bmatrix} x_k \\ y_k \\ z_k  \end{bmatrix}} &= 
\scalemath{.85}{
{^{C_k}\mathbf p_{f_i}}  
= \mathbf C(^C_I \bar{q}) \mathbf C(^{I_k}_G \bar{q}) \left({^G\mathbf p_{f_i}} - {^G\mathbf p_{I_k}} \right) + {^C\mathbf p_I}
}
 \label{eq:meas-eq}
\end{align}
where $\mathbf n_{f_k}$ is the zero-mean, white Gaussian measurement noise with covariance $\mathbf R_{k}$.
In~\eqref{eq:meas-eq}, $\{^C_I \bar{q},  {^C\mathbf p_I} \}$ 
is the extrinsic rotation and translation between the camera and IMU.
This transformation can be obtained, e.g.,
by performing camera-IMU extrinsic calibration offline~\cite{Mirzaei2008TRO}.
For the use of EKF, linearization of~\eqref{eq:meas-model}  yields 
the following residual:
\begin{align} 
 \mathbf r_{f_k}
&= \mathbf H_{k} \widetilde{\mathbf x}_{{k|k-1}} + \mathbf n_{f_k}  \\
&= \mathbf H_{I_{k}} \widetilde{\mathbf x}_{I_{k|k-1}} + \mathbf H_{f_{k}} {^G\widetilde{\mathbf p}_{f_{i,k|k-1}}} + \mathbf n_{f_k} 
\label{eq:residual}
\end{align}
where $\mathbf H_{k}$ is computed by (for simplicity assuming $i=1$):
\begingroup \setlength\arraycolsep{4pt}
\begin{align} \label{eq:meas-jac}
&\scalemath{0.9}{\mathbf H_{k} = \begin{bmatrix} \mathbf H_{I_{k}} &  \mathbf 0_{3\times 6m} &  \mathbf H_{f_{k}} &  \mathbf 0_{3\times (3n-3)}  \end{bmatrix} =
}\\
&\scalemath{.85}{
\mathbf {H_{proj}} \mathbf C(^C_I \bar{q}) \begin{bmatrix} \mathbf H_{\bm\theta_k} & \mathbf 0_{3\times 9} & \mathbf H_{\mathbf p_k}  & \mathbf 0_{3\times 6m} &   \mathbf C(^{I_k}_G \hat{\bar{q}}) &   \mathbf 0_{3\times (3n-3)}  \end{bmatrix} } \notag \\
&\scalemath{0.9}{
\mathbf {H_{proj}} = \frac{1}{\hat z_k^2} \begin{bmatrix} \hat z_k & 0 & -\hat x_k \\ 0 & \hat z_k & -\hat y_k \end{bmatrix}  
}\\
&\scalemath{.9}{
\mathbf H_{\bm\theta_k} = \lfloor  \mathbf C(^{I_k}_G \hat{\bar{q}}) \left({^G\hat{\mathbf p}_{f_i}} - {^G\hat{\mathbf p}_{I_k}} \right) \times  \rfloor  ~,~
\mathbf H_{\mathbf p_k} = -\mathbf C(^{I_k}_G \hat{\bar{q}}) 
}
\label{eq:meas-jac2}
\end{align}
\endgroup
Once the measurement Jacobian and residual are computed, 
we can apply the standard EKF update equations to update the state estimates and error covariance~\cite{Maybeck1979}.

 \section{Schmidt-EKF based VI-SLAM} \label{sec:svis}

It is known that the EKF update of state estimates and covariance has quadratic complexity in terms of the number of map features~\cite{Paz2008TRO},
making naive implementations of VI-SLAM too expensive to run in real-time.
Leveraging the SKF \cite{Schmidt1966ACS}, we propose a novel {\bf S}chmidt-{\bf E}KF for {\bf VI}-{\bf S}LAM (SEVIS) algorithm which mitigates this quadratic complexity.
The key idea is to selectively treat map features as nuisance parameters in the state vector [i.e., Schmidt state $\mathbf x_S$~\eqref{eq:state}] 
whose mean and covariance will no longer be updated,
while their cross-correlations with the active state $\mathbf x_A$ are still utilized and updated.

In particular, the IMU propagation of the proposed SEVIS is identical to that of the standard EKF in Section~\ref{sec:propagation}.
In what follows we primarily focus on the update with monocular images, which is at the core of our SEVIS, but the approach is easily extendable to stereo systems.
As the camera-IMU sensor pair moves through the environment, features are tracked using descriptor-based tracking.
FAST features are first detected~\cite{Rosten2010TPAMI} and ORB descriptors \cite{Rublee2011ICCV} are extracted for each.
The OpenCV \cite{OPENCV_library} ``BruteForce-Hamming'' KNN descriptor matcher is used to find correspondences, after which we perform both a ratio test between the top two returns to ensure valid matches and 8-point RANSAC to reject any additional outliers.
Once visual tracks are found, 
three types of tracked features are used to efficiently update state estimates and covariance:
(i) VIO features that are opportunistic and can only be tracked for a short period time,
(ii) SLAM features that are more stable than the above one and can be tracked beyond the current sliding window,
and (iii) map features that are the matured and informative SLAM features which are kept in the Schmidt state for an indefinite period of time.

\subsection{VIO Features: MSCKF Update} \label{sec:vio_feat_update}

For those features that have lost active track in the current window (termed VIO features),
we perform the standard MSCKF update~\cite{Mourikis2007ICRA}.
In particular, we first perform BA to triangulate these features for computing the feature Jacobians $\mathbf H_f$ [see \eqref{eq:meas-jac}], 
and then project $\mathbf{r}_k$ [see \eqref{eq:residual}] onto the left nullspace of $\mathbf{H}_f$ (i.e., $\mathbf{N}^\top \mathbf H_f = \mathbf 0$) to yield the measurement residual independent of features:
\begin{align} \label{eq:msckf-null-trick}
    \mathbf{N}^\top\mathbf{r}_{f} &= \mathbf{N}^\top\mathbf{H}_x\tilde{\mathbf{x}}_{A_{k|k-1}} + \mathbf{N}^\top\mathbf{H}_f{}^G\tilde{\mathbf{p}}_{f_{i}} + \mathbf{N}^\top\mathbf{n}_f \\
   \Rightarrow~ \mathbf{r}'_f &= \mathbf{H}_x'\tilde{\mathbf{x}}_{A_{k|k-1}} + \mathbf{n}_f'
   \label{eq:feat_update_linearized}
\end{align}
where $\mathbf H_x$ is the stacked measurement Jacobians with respect to the navigation states in the current sliding window,
$\mathbf R_f^\prime = \mathbf N^\top \mathbf R_f \mathbf N$ is the inferred noise covariance~\cite{Mourikis2007ICRA}.

\subsection{SLAM Features: EKF Update} \label{sec:slam_feat_update}

For those features that can be reliably tracked longer than the current sliding window,
we will initialize them into the active state and perform EKF updates as in the standard EKF-based VI-SLAM (see Section~\ref{sec:update}).
However, it should be noted that SLAM features will not remain active forever, 
instead they will either be moved to the Schmidt state as nuisance parameters (see Section~\ref{sec:schmidt_feat_update}) or marginalized out for computational savings as in~\cite{Li2012RSS}.

\subsection{Map Features: Schmidt-EKF Update} \label{sec:schmidt_feat_update}

If we perform VIO by linearly marginalizing out features~\cite{Yang2017IROS} as in the MSCKF~\cite{Mourikis2007ICRA},
the navigation errors may grow unbounded albeit achieving efficiency;
on the other hand, if performing full VI-SLAM by continuously maintaining features (map) in the state, the computational cost may become prohibitive albeit gaining accuracy.
In particular, two challenges arise in SLAM that must be tackled:
(i) the increase in computational complexity due to number of map features included, 
and (ii) the data association of detecting whether actively tracked features match previously mapped features in the state vector.
This motivates us to design our SEVIS algorithm that builds a sparse feature-based map of the environment which can then be leveraged to prevent long-term drift while still preserving  necessary efficiency via the SKF.

\subsubsection{Keyframe-aided 2D-to-2D Matching}

\begin{figure}\centering
\includegraphics[width=.7\columnwidth]{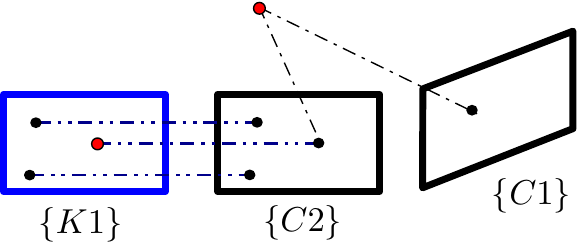}
\caption{Illustration of the proposed keyframe-aided 2D-to-2D matching for data association. 
Assuming a cloned frame $\{C2\}$ matches to a keyframe $\{K1\}$ with all actively tracked features,
and among these positive matches, one feature (red) corresponds to a map feature, 
the measurements in $\{C2\}$ and $\{C1\}$ will be used to update the active state by performing Schmidt-EKF update.
}
\label{fig:diagram_pt_meas}
\vspace{-1em}
\end{figure}

To overcome the data association challenge,
given 3D positions of map features already included in the state vector, 
one straightforward approach might be through 3D-to-2D projection (i.e., projecting the 3D map feature onto the current frame) to find the correspondence of current visual measurements to the mapped feature,
which is often used in the literature (e.g., \cite{Dutoit2017ICRA,Lynen2015RSS}).
However, in a typical SLAM scenario, estimating a map of 3D point features and matching them to current features is often sparse;
for example, we found that it was common for a multi-floor indoor environment with up to 600 map features to only have about 10 features that can successfully project back into the active frame.
Moreover, if there is any non-negligible drift in the state estimates (which is inevitable in practice), then projected features are likely to not correspond to the same spacial area as the current image is observing,
thus preventing utilization of map information to reduce navigation errors.

For these reasons, 
we advocate 2D-to-2D matching for data association with the aid of ``keyframes'' that observe previous areas in the environment,
due to its ability to provide high quality estimates and not be effected by estimation drift.
Each keyframe contains a subset of the extracted features that correspond to map features in the state vector, 
and thus, if we match active feature tracks to previous keyframes we can find the correspondence between the newly tracked features and the previously mapped features that reside in our state.

Specifically, we first query the keyframe database to retrieve the closest keyframe to the current frame.
To this end, different place recognition approaches such as DBoW2~\cite{Galvez2012TRO} and CALC~\cite{Merrill2018RSS} can be used to find the best candidate.
After retrieval, we perform an additional geometric check by ensuring that the fundamental matrix can be calculated between the current frame and the proposed keyframe match, which we found provided extremely good matches to the best keyframe in the database.
After retrieving a matching keyframe, we perform descriptor-based matching from features in the current frame to the keyframe with \textit{all} extracted features from both frames followed by 8-point RANSAC to reject outliers.
We now have the correspondences between the current frame feature tracks and keyframe map features.
Fig.~\ref{fig:diagram_pt_meas} visualizes this process.

\subsubsection{Schmidt-EKF Update} \label{sec:ekf_schmidt_update}

To gain significant computational savings while still performing SLAM and exploiting map constraints to bound navigation errors,
we adapt the SKF methodology~\cite{Schmidt1966ACS} and treat the map features as nuisance parameters by only tracking their cross-correlations with the active states while still allowing for probabilistic inclusion of them during update.
In particular, we compute the gain matrix of the Schmidt-EKF as follows:
\begin{align}
\scalemath{1}{
\begin{bmatrix} \mathbf K_{A_k} \\ \mathbf K_{S_k} \end{bmatrix} 
=  \begin{bmatrix}     
\mathbf{P}_{AA_{k|k-1}} \mathbf H_{A_k}^\top + \mathbf{P}_{AS_{k|k-1}}\mathbf H_{S_k}^\top \\
 \mathbf{P}_{SA_{k|k-1}} \mathbf H_{A_k}^\top + \mathbf{P}_{SS_{k|k-1}} \mathbf H_{S_k}^\top     
 \end{bmatrix} 
\mathbf S_k^{-1} 
}
\label{eq:kalman-gain}
\end{align}
where $\mathbf H_A$ and $\mathbf H_S$ are respectively the measurement Jacobians with respect to the active and Schmidt states features [see~\eqref{eq:residual}],
and $\mathbf{S}_k$ is the residual covariance given by:
\begin{equation}
    \mathbf{S}_k = \begin{bmatrix} \mathbf{H}_{A_k} & \mathbf{H}_{S_k}\end{bmatrix} \mathbf{P}_{k|k-1}
    \begin{bmatrix} \mathbf{H}_{A_k} & \mathbf{H}_{S_k}\end{bmatrix}^{\top} 
    + \mathbf{R}_k
    \label{eq:schmidt_SK}
\end{equation}
To reduce the computational complexity, we do {\em not} update the map feature nuisance parameters (Schmidt state), 
and thus, as in the SKF, we set the gain corresponding to the Schmidt state to zero, i.e., $\mathbf K_{S_k} = \mathbf 0$.
As a result, the state estimate is updated as follows:
\begin{align}
    \hat{\mathbf x}_{A_{k|k}} = \hat{\mathbf x}_{A_{k|k-1}}+\mathbf K_{A_k} \mathbf r_{f_k}~,~~~
   \hat{ \mathbf x}_{S_{k|k}} = \hat{\mathbf x}_{S_{k|k-1}}
    \label{eq:schmidt_update}
\end{align}
The covariance is efficiently updated in its partitioned form:
\begin{align}
    \mathbf{P}_{AA_{k|k}} &= \mathbf{P}_{AA_{k|k-1}} \notag\\
    &- \mathbf{K}_{A_k}(\mathbf{H}_{A_k}\mathbf{P}_{AA_{k|k-1}}
    +\mathbf{H}_{S_k}\mathbf{P}_{AS_{k|k-1}}^{\top}) \label{eq:covAA_update}\\
    \mathbf{P}_{AS_{k|k}} &= \mathbf{P}_{AS_{k|k-1}}  \notag\\
    &- \mathbf{K}_{A_k}(\mathbf{H}_{A_k}\mathbf{P}_{AS_{k|k-1}}
    +\mathbf{H}_{S_k}\mathbf{P}_{SS_{k|k-1}}) \\
    \mathbf{P}_{SS_{k|k}} &= \mathbf{P}_{SS_{k|k-1}}
\end{align}
Up to this point, we have fully utilized the current camera measurement information to update the SEVIS state estimates and covariance [see~\eqref{eq:state} and \eqref{eq:cov}].
The main steps of the proposed SEVIS are outlined in Algorithm~\ref{alg:sevis}.

\begin{algorithm}
\caption{Schimdt-EKF Visual-Inertial SLAM (SEVIS)}
\begin{algorithmic}\STATE \textbf{Propagation}: Propagate the IMU navigation state estimate $\hat{\mathbf x}_{I_{k|k-1}}$ based on~\eqref{eq:imu_dynamics}, 
the active state's covariance $\mathbf{P}_{AA_{k|k-1}}$ and cross-correlation $\mathbf{P}_{AS_{k|k-1}}$ based on~\eqref{eq:propcov}.
\STATE \textbf{Update}: For an incoming image,
\STATE
\begin{itemize}[nolistsep]
    \item Perform stochastic cloning~\cite{Roumeliotis2002ICRAa} of current state.
    \item Track features into the newest frame.
    \item Perform keyframe-aided 2D-to-2D matching to find map feature correspondences:
    \begin{itemize}[nolistsep]
        \item Query keyframe database for a keyframe visually similar to current frame.
        \item Match currently active features to the features in the keyframe.
        \item Associate those active features with mapped features in the keyframe.
    \end{itemize}
    \item Perform MSCKF update for VIO features (i.e., those that have lost their tracks) as in Section~\ref{sec:vio_feat_update}.
    \item Initialize new SLAM features if needed and perform EKF update as in Section~\ref{sec:slam_feat_update}.
    \item Perform Schmidt-EKF update for map features as in Section~\ref{sec:ekf_schmidt_update}.
\end{itemize}
\STATE \textbf{Management of Features and Keyframes}: 
\STATE
\begin{itemize}[nolistsep]
    \item Active SLAM features that have lost track are moved to the Schmidt state or marginalized out.
    \item Marginalize the oldest cloned pose from the sliding window state.
    \item Marginalize map features if exceeding the maximum map size.
    \item Insert a new keyframe into database if we have many map features in the current view.
    \item Remove keyframes without map features in view.
\end{itemize}
\end{algorithmic}
\label{alg:sevis}
\end{algorithm}

\subsection{Computational Complexity Analysis}

Here we demonstrate the computational efficiency of the proposed SEVIS by providing detailed analysis that shows the complexity is {\em linear} with respect to the number of map features.  This efficiency will also be demonstrated with experimental data in Section \ref{sec:exp}.

\noindent\textbf{Propagation}:
The main computational cost of propagation comes from the matrix multiplication of $\bm\Phi_{k-1} \mathbf{P}_{AS_{k-1|k-1}}$ [see \eqref{eq:propcov}], 
where $\bm\Phi_{k-1}$ is a square matrix of $\dim(\mathbf x_A)$ size 
and $\mathbf{P}_{AS_{k-1|k-1}}$ is a fat matrix with size of $O(n)$.
This incurs a total cost of $O(n)$
because the number of map features far exceeds the size of the active state. 

\noindent\textbf{Update}:
After propagation, we augment the state by appending the propagated state to the active clone state $\mathbf{x}_C$.
This is an $O(n)$ computation as we simply need to append a new row and column on the active covariance $\mathbf{P}_{AA_k}$ and then a row on the Schmidt cross-correlation terms $\mathbf{P}_{AS_k}$ yielding an $O(n)$ operation.
SLAM feature initialization follows the same logic and is an $O(n)$ operation.
During update, naively, the operation allows for the computation cost to be on order $O(n^2)$ in the case that the current frame matches to {\em all} features in the map at the same time instance.
A close inspection of \eqref{eq:schmidt_SK} reveals that if the size of $\mathbf{H}_{S_k}$ is order $n$, the calculation of $\mathbf{S}_k$ will be of $O(n^2)$.
However, this is {\em not} common in practice due to large environments and limited viewpoints. 
Therefore, we limit the number of features that can be used in one update to be far lower than the order of $n$ and additional features can be processed at future instances to spread the computation over a period of time allowing for $O(n)$ complexity at every time step.

\noindent\textbf{Management}:
We manage the matrices $\mathbf{P}_{AA_k}$, $\mathbf{P}_{AS_k}$, and $\mathbf{P}_{SS_k}$ as separate entities and pre-allocate $\mathbf{P}_{SS_k}$ to the maximum number of allowed features to prevent overhead from memory allocation operations.
When moving a state from the active state $\mathbf{x}_{A_k}$ to the Schmidt state $\mathbf{x}_{S_k}$, special care is taken such that this operation remains on order $O(n)$.
In particular, 
we first copy the associated block column from $\mathbf{P}_{AA_k}$ onto the last column of $\mathbf{P}_{AS_k}$, 
after which we copy the associated block row in $\mathbf{P}_{AS_k}$ to the last row and column in the pre-allocated $\mathbf{P}_{SS_k}$, thus yielding an total cost of $O(n)$.
Marginalizing states in the active state is of $O(n)$ as it requires removal of a row from the $\mathbf{P}_{AS_k}$ matrix which is achieved through copying all rows after the to-be-removed upwards overwriting the to-be-removed entries.
During marginalization of map features from the Schmidt state $\mathbf{P}_{SS_k}$, we overwrite the rows and columns corresponding with the to-be-removed state with the last inserted map feature, allowing for an $O(n)$ operation.

\section{Monte-Carlo Simulation Results} \label{sec:sim}

\begin{figure}
                \subfloat[Orientation RSSE]{\includegraphics[width = .5\columnwidth]{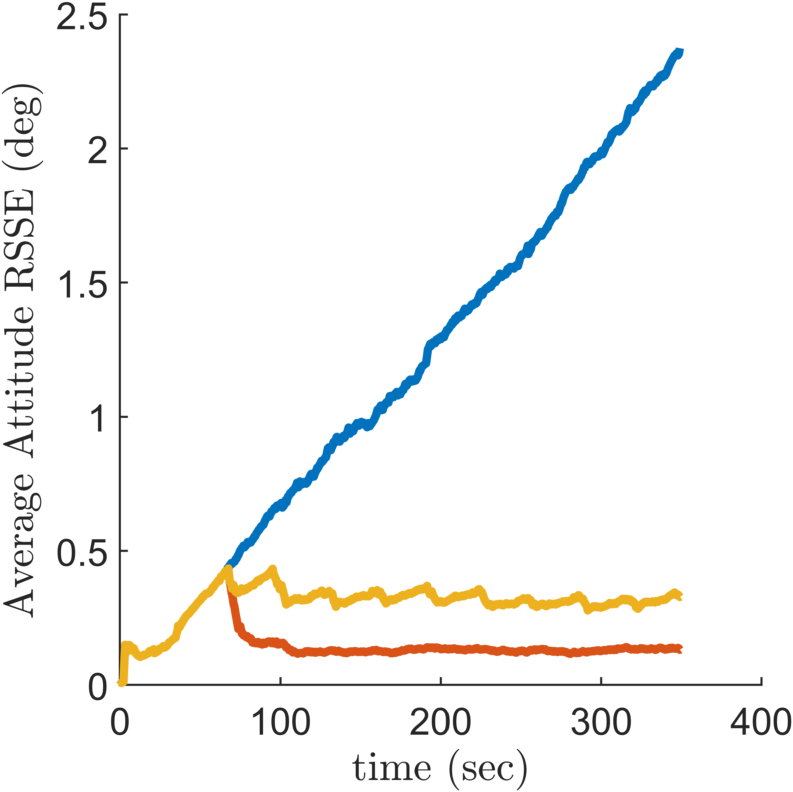} \label{fig:sim_indoor_perf_1}}
    \subfloat[Position RSSE]{\includegraphics[width = .5\columnwidth]{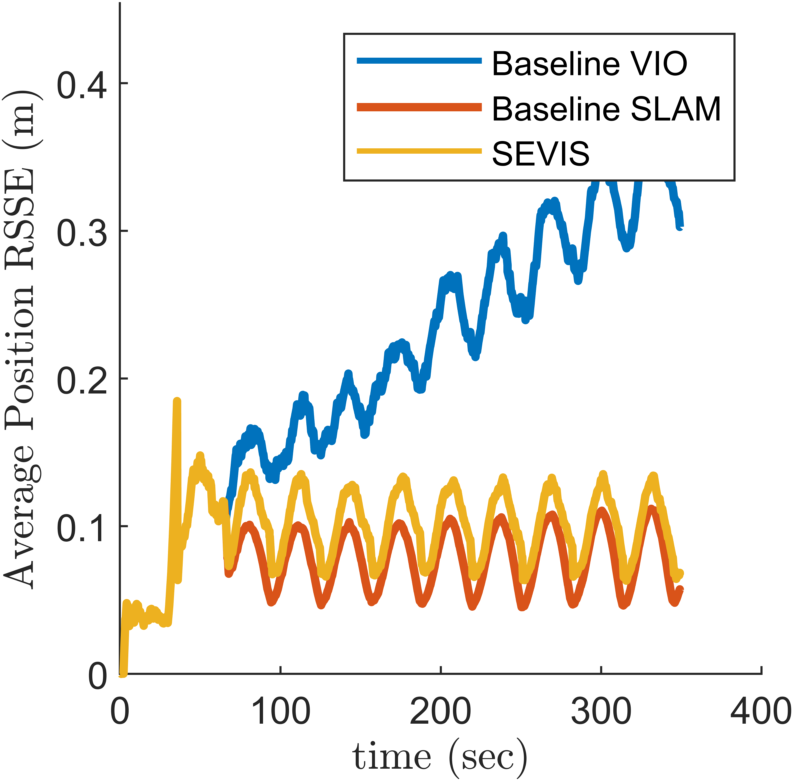}     \label{fig:sim_indoor_perf_2}}
    \caption{Monte-Carlo simulation
        averaged RSSE of pose (position and orientation) estimates for the three considered VIO and VI-SLAM algorithms.}
    \label{fig:sim_indoor_results}
\vspace{-1em}
\end{figure}

To validate the back-end estimation engine of the proposed SEVIS,
we first perform Monte-Carlo simulations of visual-inertial SLAM with known measurement-feature correspondences,
where a monocular-visual-inertial sensor platform is moving on a circular trajectory within a cylinder arena observing a series of environmental features.
The simulation  parameters about the sensors and the trajectory are listed in Table \ref{tab:simulation_parameters}.

\begin{table}\small
\caption{Monte-Carlo Simulation Parameters}
\label{tab:simulation_parameters}
\begin{tabular}{@{}lll@{}}
\toprule
\textbf{Parameter} & \textbf{Value} & \textbf{Units} \\ \midrule
IMU Angle Random Walk Coeff. & 0.4   & deg/$\sqrt{\text{Hr}}$  \\
IMU Rate Random Walk Coeff. & 0.02 & deg/sec/$\sqrt{\text{Hr}}$ \\
IMU Velocity Random Walk Coeff.     & 0.03  & m/sec/$\sqrt{\text{Hr}}$   \\ 
IMU Acceleration Random Walk Coeff.     & 0.25  & milli-G/$\sqrt{\text{Hr}}$   \\ 
IMU Sample Rate     & 100   & Hz   \\ 
Image Processing Rate & 5 & Hz \\
Feature Point Error $1\sigma$ & 0.17 & deg \\
Number of MSCKF Poses & 15 &  \\
Approximate Loop Period & 32 & sec \\
\bottomrule
\end{tabular}
\end{table}

In particular, we compare three VINS algorithms to reveal 
the benefits of the proposed SEVIS:
(i) 
The baseline VIO approach, which consists of the MSCKF augmented with 6 SLAM features (see \cite{Li2012RSS}).
These SLAM features are explicitly marginalized out when they leave the field of view.  
(ii) 
The baseline SLAM method, which uses the same MSCKF window but is augmented with 90 SLAM features.  
Different from the above VIO, in this case the SLAM features are never marginalized so that they can be used for (implicit) loop closures.  
(iii) The proposed SEVIS algorithm, which consists of the same MSCKF window and 6 SLAM features as in the baseline VIO, while being augmented with a bank of 90 map features that are modeled as nuisance parameters. 
When the SLAM features leave the field of view, they are moved into the Schmidt states, becoming the map features as described in Algorithm~\ref{alg:sevis}.

The average root sum squared error (RSSE) performance of 50 Monte-Carlo simulation runs are shown in Fig. \ref{fig:sim_indoor_results}.
As expected, the baseline VIO  accumulates drift in both orientation and position over time
while the baseline SLAM provides bounded error performance without long term drift.
It is interesting to point out that the position RSSE oscillates slightly depending on the location relative to the initial loop closure.
This is because that the EKF has limited ability to correct these errors as it cannot re-linearize past measurements unlike optimization-based approaches~\cite{Triggs2000VA}.
More importantly, it is clear that 
the proposed SEVIS algorithm also does not accumulate long-term drift, 
although it is slightly less accurate than the baseline SLAM.  
However, this degradation in accuracy is a small price to pay 
considering that the SEVIS is of {\em linear} computational complexity with respect to the number of map features, 
while the baseline SLAM has {\em quadratic} complexity.

 \section{Real-World Experimental Results}
\label{sec:exp}

We further evaluated the baseline MSCKF-based VIO (without map features), 
the baseline full VI-SLAM, and the proposed SEVIS on real-world datasets.
In what follows, we first examine the estimator accuracy and computational overhead, after which the systems are evaluated on a challenging nighttime multi-floor dataset, showing that the proposed SEVIS can robustly be extended to realistic applications.

\subsection{Vicon Loops Dataset}

\begin{figure}\centering
\includegraphics[width=0.95\columnwidth]{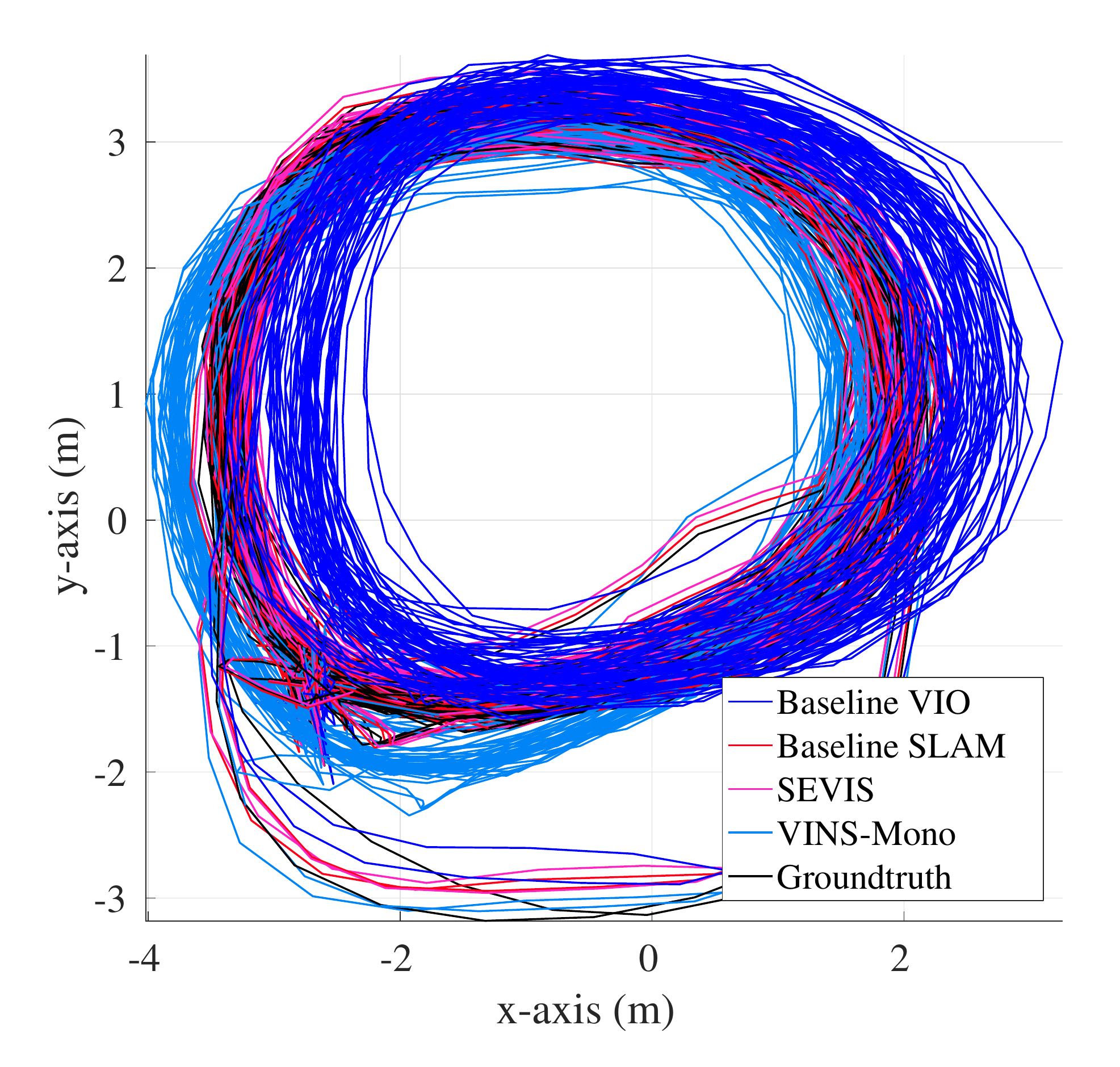}
\includegraphics[width=0.95\columnwidth]{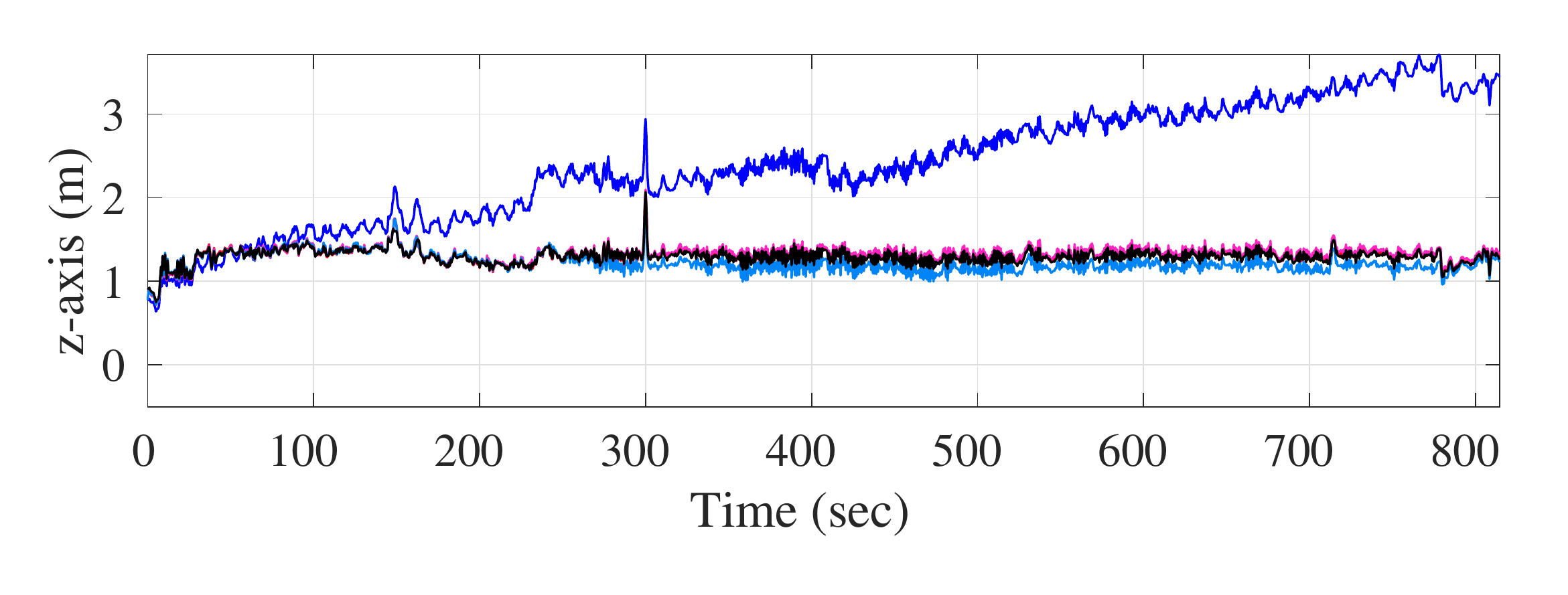}
\vspace*{-0.4cm}
\caption{
Trajectory of the baseline VIO, baseline SLAM with map features, proposed SEVIS with Schmidt covariance update, and VINS-Mono \cite{Qin2018TRO,Qin2018RELOC}.
Clearly the inclusion of map features has limited the drift and allows for high accuracy.
}
\label{fig:loop_path}
\vspace{-0.5em}
\end{figure}

\begin{table}[]
\centering
\caption{
Relative trajectory error for different segment lengths along with the overall absolute trajectory error. Values where computed using Zhang and Scaramuzza's open sourced utility \cite{Zhang2018IROS}.
}
\begin{tabular}{C{1.25cm}C{1.25cm}C{1.25cm}C{1.25cm}C{1.25cm}} \toprule
\textbf{Segment Length} & \textbf{Baseline VIO} & \textbf{Baseline SLAM} & \textbf{SEVIS} & \textbf{VINS-Mono} \\ \hline
123m & 0.383 & 0.102 & 0.111 &  0.184 \\
247m & 0.645 & 0.099 & 0.108 &  0.238 \\
370m & 0.874 & 0.104 & 0.123 &  0.325 \\
494m & 1.023 & 0.095 & 0.121 &  0.381 \\
618m & 1.173 & 0.107 & 0.139 &  0.425 \\ \hline
ATE  & 0.779 & 0.121 & 0.128 &  0.323 \\ \hline
\end{tabular}
\label{table:rmseokvis}
\end{table}

\begin{figure}\centering
\includegraphics[width=1.0\columnwidth]{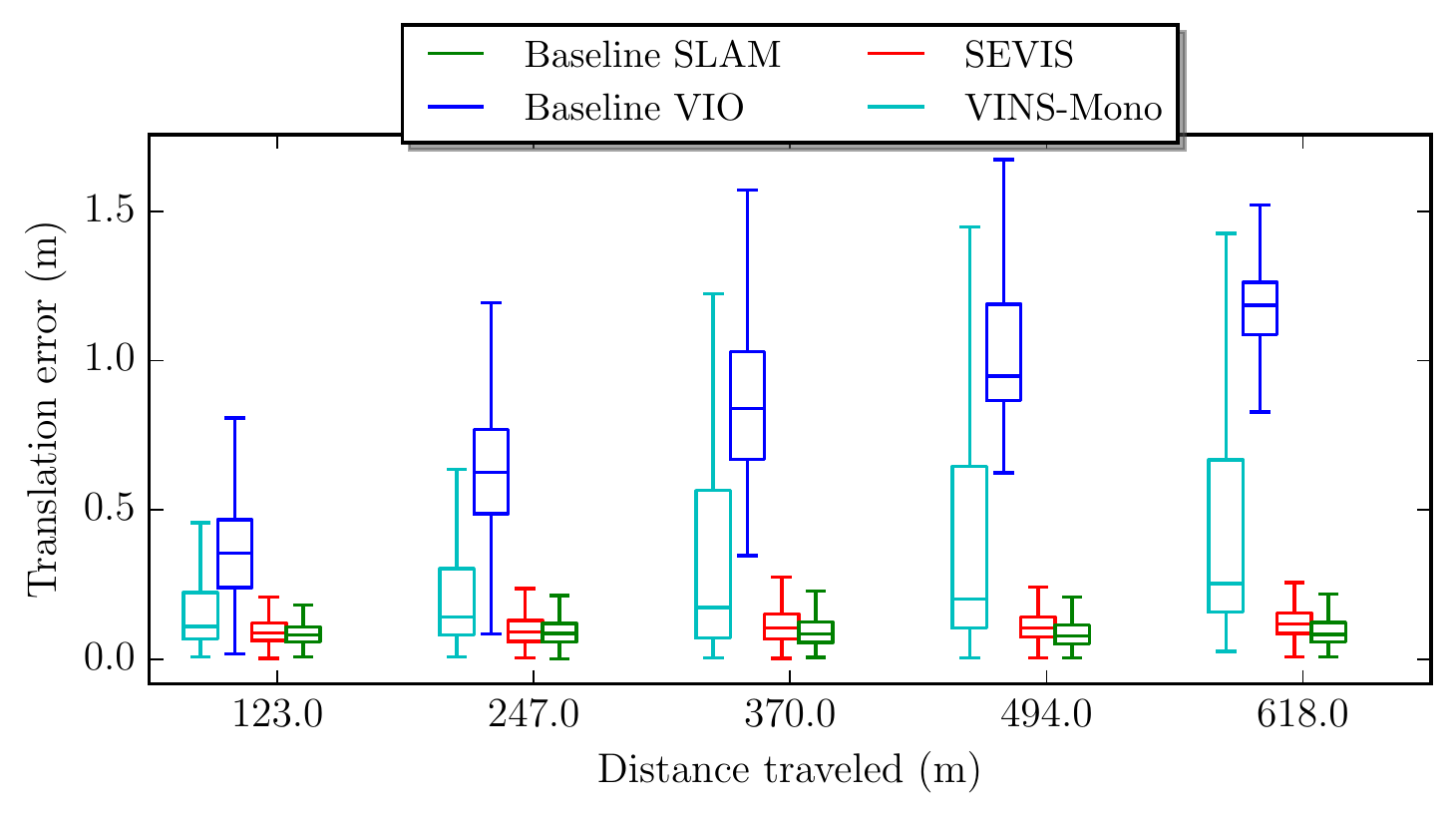}
\caption{
Boxplot of the relative trajectory error statistics.
The middle box spans the first and third quartiles, while the whiskers are the upper and lower limits. Plot best seen in color.
}
\label{fig:relposerror}
\vspace{-0.5em}
\end{figure}

\begin{figure}\centering
\hspace*{-0.8cm}
\includegraphics[width=1.0\columnwidth]{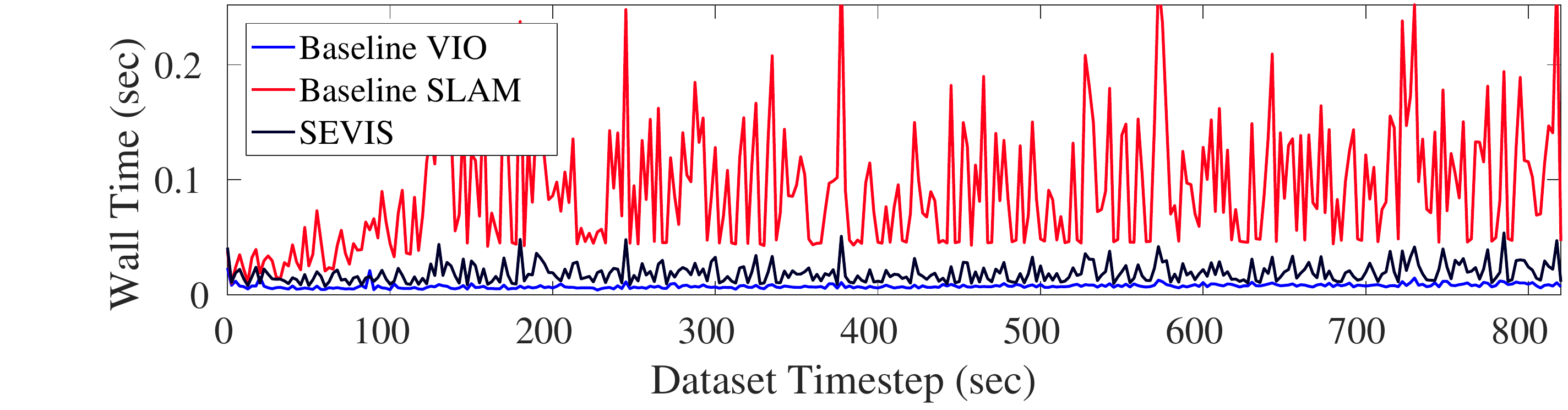}
\caption{
The wall clock execution time in seconds comparing the three methods can be seen.
Plot best seen in color.
}
\label{fig:loop_timing}
\vspace{-1em}
\end{figure}

We first validated the proposed system on the Vicon loops dataset \cite{Leutenegger2014IJRR} that spans {\bf 1.2km} in a single room over a 13 minute collection period.
A hand-held VI-sensor \cite{Nikolic2014ICRA} provides grayscale stereo image pairs and inertial information, while full 6DOF groundtruth is captured using a Vicon motion tracking system at 200 Hz.
The maximum number of map features was set to 600 points to ensure real-time performance over the entire trajectory with images inserted into the query keyframe database at 0.5 Hz and a max of 5 SLAM features in the active state at a time.
The results presented show three different configurations: (i) the baseline VIO augmented with 5 SLAM features, (ii) the baseline VI-SLAM with 600 SLAM/map features, and (iii) the proposed SEVIS with 600 map features that leverages the Schmidt formulation for computational gains.

We evaluated the proposed method using two different error metrics: Absolute Trajectory Error (ATE) and Relative Error (RE).
We point the reader to \cite{Zhang2018IROS} for detailed definitions of these error metrics.
Alongside our baseline and proposed methods, we additionally evaluated VINS-Mono~\cite{Qin2018TRO,Qin2018RELOC} to provide a comparison to a current state-of-the-art method that leverages loop closure information.
Shown in Table \ref{table:rmseokvis} and Fig. \ref{fig:relposerror}, the proposed SEVIS is able to localize with high accuracy and perform on the level of the full baseline VI-SLAM system.
Looking at the RE it is clear that the inclusion of map features prevents long-term drift and offers a greater accuracy shown by the almost constant RE as the trajectory segment length grows.
The proposed SEVIS provides a computationally feasible filter that has similar accuracy as full baseline VI-SLAM with competitive performance to that of VINS-Mono (although the VINS-Mono leverages batch optimization).

The primary advantage of the proposed SEVIS algorithm over full-covariance SLAM is a decrease in computational complexity.
The practical utility of this is evident in the run-times of the different algorithms.
As shown in Fig.~\ref{fig:loop_timing}, we evaluated the three systems and collected timing statistics of our implementation.\footnote{Single thread on an Intel(R) Xeon(R) E3-1505Mv6 @ 3.00GHz}
The proposed SEVIS is able to remain real-time (20 Hz camera means we need to be under 0.05 seconds total computation), while the full VI-SLAM method with 600 map features, has update spikes that reach magnitudes greater then four times the computational limit.
This is due to the full covariance update 
being of order $O(n^2)$.
Note that there is an additional overhead in the propagation stage as symmetry of the covariance matrix needs to be enforced for the entire matrix instead of just the active elements to ensure numerical stability.

\subsection{Nighttime Multi-Floor Dataset}

\begin{figure}\centering
\includegraphics[width=0.85\columnwidth]{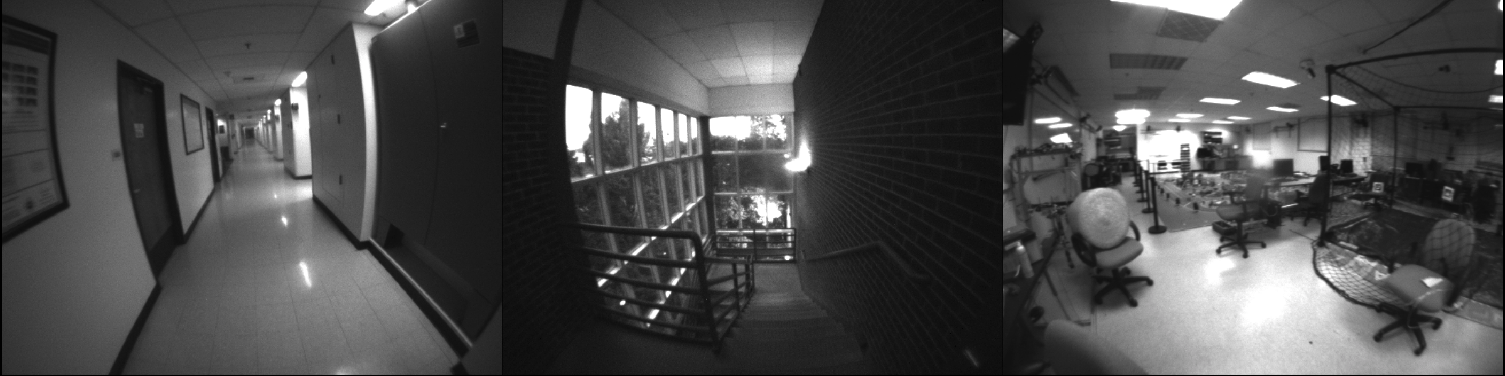}
\caption{Selected views during the night multi-floor trajectory  show the high noise, poor lighting conditions, and motion blur that greatly challenge visual feature tracking.
}
\label{fig:spencer_imgs}
\vspace{-1em}
\end{figure}

We further challenged the proposed system on a difficult indoor nighttime multi-floor dataset,
which has multiple challenges including
low light environments, long exposure times, and low contrast images with motion blur unsuitable for proper feature extraction (see Fig.~\ref{fig:spencer_imgs}).
If features can be extracted, the resulting descriptor matching is poor due to the high noise and small gradients, and as compared to the Vicon Loops Dataset, more outliers are used during update, causing large estimator jumps and incorrect corrections.
We stress that the proposed SEVIS can recover in these scenarios due to keyframe-aided 2D-to-2D matches which are invariant to poor estimator performance or drift and map feature updates correct and prevent incorrect drift.

\begin{figure}\centering
\includegraphics[width=.85\columnwidth]{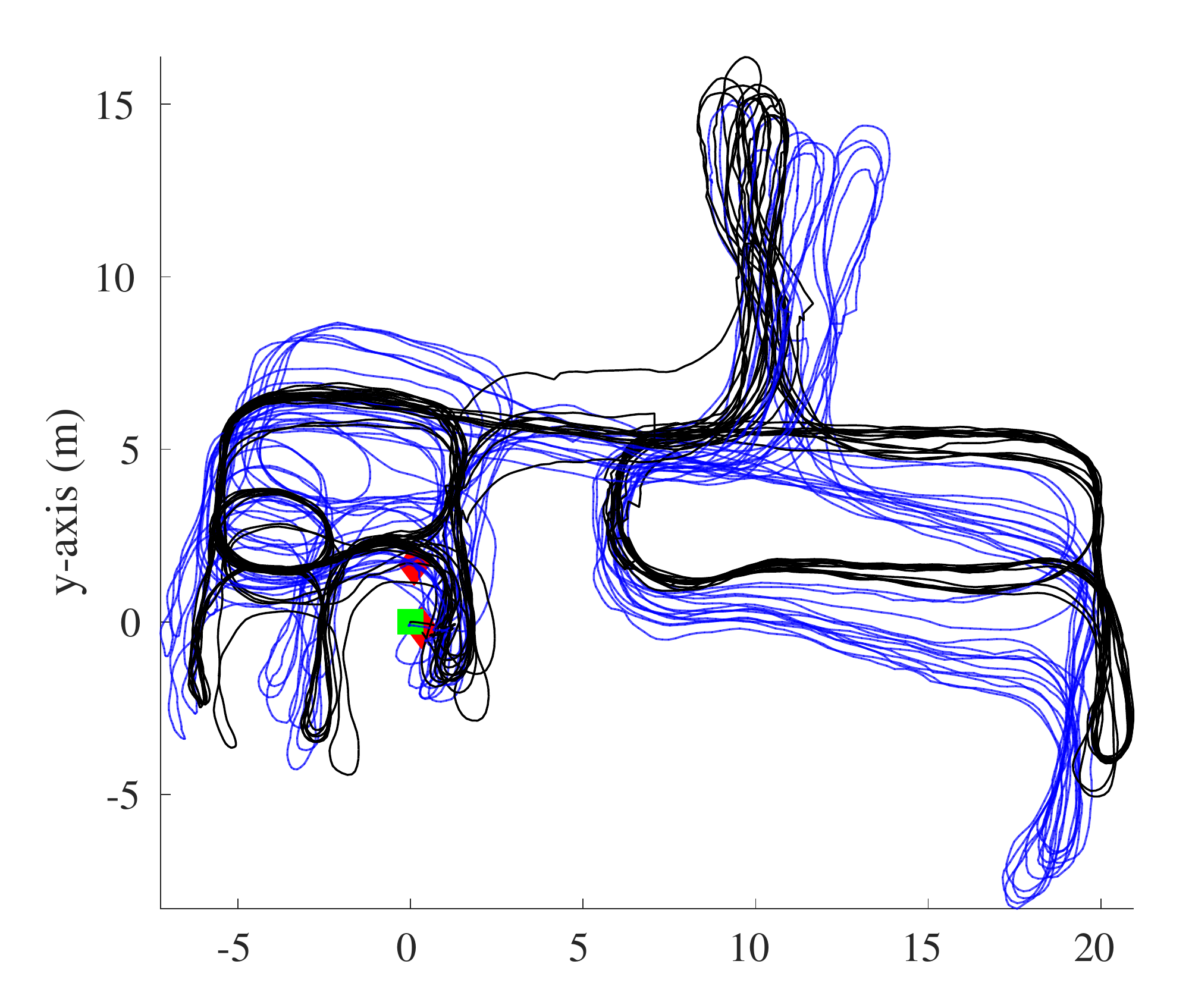}
\includegraphics[width=.85\columnwidth]{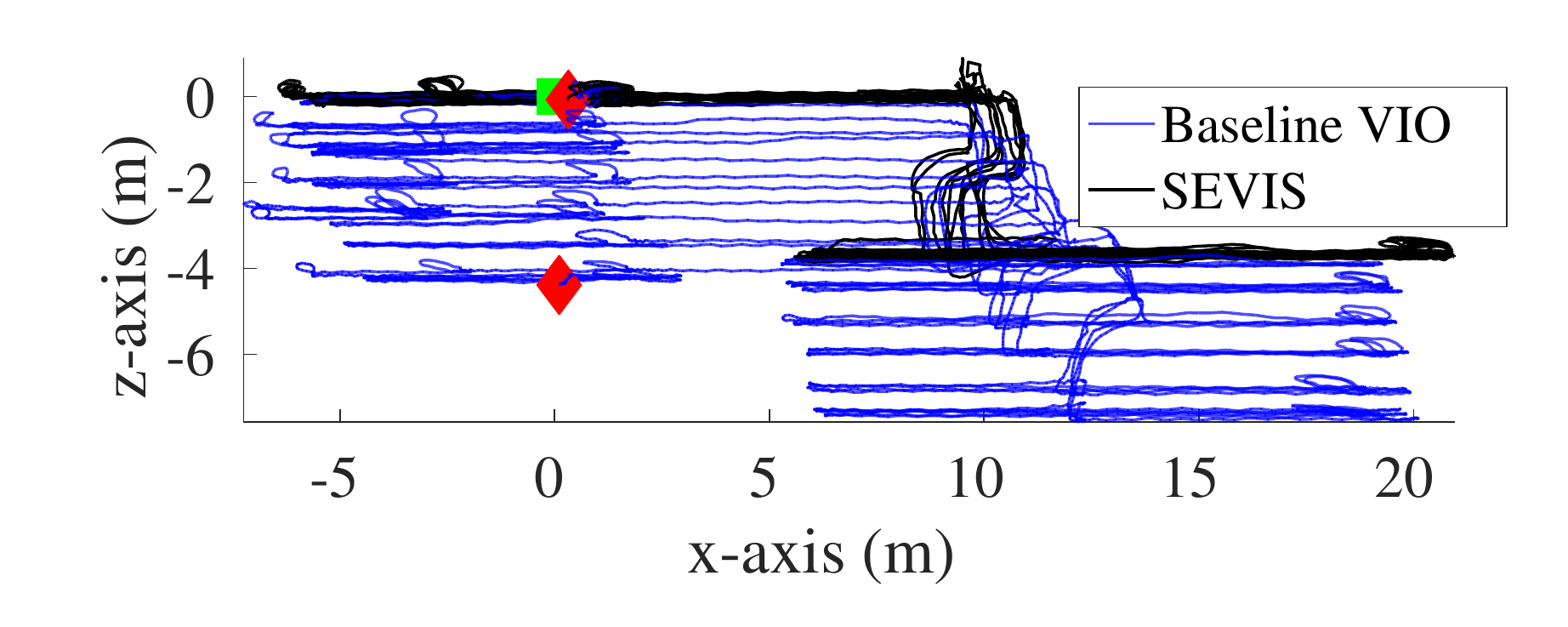}
\caption{
Estimated trajectories of the baseline VIO (blue) and SEVIS (black) show the improved performance due to inclusion of map features.
The start-end positions are denoted with a green square and red diamond respectively.
}
\label{fig:spencer_path}
\vspace{-1.5em}
\end{figure}

A Realsense ZR300 sensor\footnote{\href{https://software.intel.com/en-us/realsense/zr300}{https://software.intel.com/en-us/realsense/zr300}} was used to collect 20 minutes of grayscaled monocular fisheye images with inertial readings, with the {\bf 1.5km} trajectory spanning two floors.
We additionally performed online calibration of the camera to IMU extrinsic to further refine the transform provided by the manufacture's driver.
A max of 700 map-points allowed for sufficient coverage of the mapping area, keyframes where inserted into the query database at 4Hz to ensure sufficient coverage of all map features, and 2 SLAM features in the active state at a time.
The trajectory generated by the baseline VIO and the proposed SEVIS are shown in Fig.~\ref{fig:spencer_path}.
Clearly, the inclusion of map features prevent long-term drift experienced by the baseline VIO which exhibits large errors in both the yaw and z-axis direction.
Since no groundtruth was available for this dataset, as a common practice, we computed the start-end error of the trajectory which should ideally be equal to zero as the sensor platform was returned to the starting location.
The baseline VIO  had an error of {\bf 4.67m}  ({\bf 0.31\%} of trajectory distance) 
while the proposed SEVIS had an error of only {\bf 0.37m} ({\bf 0.02\%} of trajectory distance).


 \section{Conclusions and Future Work}

In this paper, we have developed the high-precision, efficient SEVIS algorithm that adapts the SKF formulation for long-term visual-inertial SLAM.
In particular, the probabilistic inclusion of map features within SEVIS allows for bounded navigation drifts while retaining linear computational complexity.
To achieves this, the keyframe-aided 2D-to-2D feature matching of current visual measurements to 3D map features greatly facilitates the full utilization of the map information.
We then performed extensive Monte-Carlo simulations and real-world experiments 
whose results showed that the inclusion of map features greatly impact the long-term accuracy while the proposed SEVIS still allows for real-time performance without effecting estimator performance.
In the future, we will investigate how to refine the quality of map features added for long-term localization 
and further evaluate our system on resource-constrained mobile sensor systems.

\section{Acknowledgment}

This work was partially supported by the University of Delaware (UD)
College of Engineering, Google Daydream, and by the U.S. Army Research Lab.

{
\bibliographystyle{packages/ieee}
\bibliography{library/extra,library/rpng,library/vins}
}

\end{document}